# $\mathcal{CSWA}$: Aggregation-Free Spatial-Temporal Community Sensing

**Jiang Bian***, **Haoyi Xiong***, **Yanjie Fu, and Sajal K. Das**
Missouri University of Science and Technology, United States
*Equal Contribution

## Abstract

In this paper, we present a novel community sensing paradigm $\mathcal{CSWA}$ –*Community Sensing Without Sensor/Location Data Aggregation*. $\mathcal{CSWA}$ is designed to obtain the environment information (e.g., air pollution or temperature) in each subarea of the target area, without aggregating sensor and location data collected by community members. $\mathcal{CSWA}$ operates on top of a secured peer-to-peer network over the community members and proposes a novel *Decentralized Spatial-Temporal Compressive Sensing* framework based on *Parallelized Stochastic Gradient Descent*. Through learning the *low-rank structure* via distributed optimization, $\mathcal{CSWA}$ approximates the value of the sensor data in each subarea (both covered and uncovered) for each sensing cycle using the sensor data locally stored in each member's mobile device. Simulation experiments based on real-world datasets demonstrate that $\mathcal{CSWA}$ exhibits low approximation error (i.e., less than $0.2°C$ in city-wide temperature sensing task and 10 units of PM2.5 index in urban air pollution sensing) and performs comparably to (sometimes better than) state-of-the-art algorithms based on the data aggregation and centralized computation.

## Introduction

Spatial-temporal community sensing is an efficient paradigm that leverages the mobile sensors of community members to monitor the spatial-temporal phenomena in the environment, such as air pollution or temperature. According to (Zhang et al. 2014a), there are two major roles in community sensing – the organizer and the participants – where the former is the individual or organization that creates the sensing task, recruits participants and collects the sensor data, while the latter (i.e., participants) involve in the sensing task and provide the sensing data. Frequently, the organizer pursues a high (or even full) spatial-temporal coverage of the collected sensor data. However incentives (e.g., monetary rewards) and the threats to privacy (e.g., exposing real-time locations) are two major concerns that may affect the willingness of the participants to join a community sensing task.

In addition to the community sensing paradigm, a wide-spectrum of applications, ranging from vehicle traffic monitoring (Ji, Zheng, and Li 2016; Yoon, Noble, and Liu 2007; Herring et al. 2010; Hachem, Pathak, and Issarny 2013; Pan et al. 2013) to air quality sensing (Aberer et al. 2010) and urban noise monitoring (Liu et al. 2014), have been proposed to efficiently monitor the environment of a large area through aggregating the real-time sensor and location data from the participants. Such applications use spatial-temporal coverage as the metric for overall task performance. Specifically, to characterize spatial-temporal coverage, the target area is split into subareas and the sensing duration is divided into a sequence of equal-length sensing cycles. In this way, the fraction of subareas covered by at least one sensor reading in each cycle is used to measure the spatial-temporal coverage.

For example, (Xiong et al. 2015; Sheng, Tang, and Zhang 2012) proposed to use the *full spatial-temporal coverage* as the criterion of the participants selection for community sensing, while (Zhang et al. 2014b; Hachem, Pathak, and Issarny 2013) studied the *partial spatial-temporal coverage* as the objectives of the optimization for budget-constrained participant selection. With the sensor data that partially covers the target area, (Wang et al. 2015; 2016a; 2017b) proposed *compressive community sensing*, which is capable of recovering the missing sensor data of the uncovered subareas from the data collected. Through the compressive community sensing, it is possible to accurately monitor the target area with even lower spatial-temporal coverage, thus resulting in reduced incentive consumption and fewer participants involved.

Though compressive community sensing can effectively reduce the required incentives and participants, it still aggregates the real-time location and sensor data from each participant, so as to first identify the covered subareas, fill with collected data, and then recover the missing data for the rest. To protect the location privacy of participants, the same of group of researchers (Wang et al. 2017a; 2016b) proposed to leverage the *Differential Geo-Obfuscation* to replace each participants' real-time location with a "mock" location while insuring the recovery accuracy. With the Differential Geo-Obfuscation, the participants' locations are expected to be well obfuscated; however, it is still possible to attack the participant's location when certain prior knowledge is leaked. Thus, in our research, to further protect the real-time location privacy, we propose a novel *Aggregation-Free* Compressive Community Sensing framework, with following assumptions:

- **Assumption I:** *the organizer is **NOT** allowed to collect the real-time location or the sensor data from any participant*;

- **Assumption II:** *Each participant covers one or multiple subareas in each sensing cycles with his/her mobility, while the location and sensor data is locally stored on his/her mobile device without raw location/sensor data sharing.*

Thus, to achieve the above goals, we propose a novel community sensing paradigm $\mathcal{CSWA}$. Specifically, $\mathcal{CSWA}$ first establishes secured peer-to-peer network connections between each pairs participants. Then, $\mathcal{CSWA}$ proposes a decentralized non-negative matrix factorization algorithm based on *Parallelized Stochastic Gradient Descent* framework. Through learning the low-rank structure via distributed optimization, $\mathcal{CSWA}$ approximates the value of sensor data in each subarea (both covered and uncovered) for each sensing cycle using the sensor data that are locally stored in each participant's mobile device.

The contributions of this paper are as follows:

- We propose a novel community sensing framework $\mathcal{CSWA}$, which is used to recover the environmental information in subareas, without aggregating sensor and location data from the community members who partially cover the target area. To the best of our knowledge, this paper is the first work that studies the problem of aggregation-free community sensing, by addressing the location privacy, distributed computing and optimization issues.

- To enable community sensing without location/sensor data aggregation, $\mathcal{CSWA}$ proposes a novel decentralized spatial-temporal compressive sensing framework that recovers the spatial-temporal information via decentralized Non-negative Matrix Factorization (NMF).The proposed solution operates on top of the *parallelized stochastic gradient descent*, which minimizes the loss function of NMF through secure Peer-to-Peer (P2P) message-passing over community members. The algorithm analysis shows that the proposed solution can efficiently approximates to the centralized NMF with the tolerable worst-case communication complexity.

- We evaluate $\mathcal{CSWA}$ using two large real-world datasets (i.e., temperature and air pollution). The experimental results demonstrate that $\mathcal{CSWA}$ tightly approximates to the state-of-the-art algorithms based on the data aggregation with centralized computation, and it outperforms the rest baselines with significant margin.

The rest of the paper is organized as follows. In the Preliminaries and Problem formulation section, we review the compressive community sensing and the matrix factorization approach. Then we introduce the parallelized stochastic gradient descent and present the problem formulation. In the Frameworks and Algorithms section, we propose framework of $\mathcal{CSWA}$ and present the algorithms in details. In the Experiments section, we evaluate $\mathcal{CSWA}$ on real-world datasets and compare it with baseline algorithms. Finally, in the Conclusion section, we summarize the work in this paper.

## Preliminaries and Problem Formulation

In this section, we first briefly introduce the previous work on the compressive community sensing. Then, we formulate the problem of our research.

**Compressive Community Sensing**

To derive the target full sensing matrix from partially collected sensing readings, the compressive community sensing (Wang et al. 2015; 2016a) is commonly considered to be an effective approach, which consists of two parts:

*Sensing Data Aggregation –* Given the target region splitting into a set of subareas (denoted as $S$) and a set of $m$ participants, in order to obtain the full picture of the target region for each sensing cycle (e.g., the $t^{th}$ cycle), the Compressive Community Sensing system first collects the sensing data from all participants. Specifically, the subareas covered by the $j^{th}$ participant in the $t^{th}$ sensing cycle ($t \in T$) is denoted as $S_j^t \subset S$. Thus, the overall coverage in the sensing cycle $t$ can be denoted as $S^t = S_1^t \cup S_2^t \cup ... \cup S_m^t$. Due to the limited mobility of each participant and limited number of participants involved, the overall coverage can usually include a subset of subareas, i.e., $S^t \subseteq S$. Given the collected sensing data, the compressive community sensing system aggregates the data and assigns each covered subarea an unique sensor data value. For example, if multiple sensor data values are collected (from multiple participants) that covers the same subarea in a sensing cycle, the *averaged* value would be used as the value of such subarea in the sensing cycle. In this way, each subarea $s \in S^t$ has been covered with one sensor data value, through data aggregation, and the compressive community sensing system needs to infer the missing sensor data of the subareas in $S \backslash S^t$ to obtain the sensor data of the whole target area.

*Missing Data Inference –* Given the aggregated sensor data of the covered subareas ($S^t$), there exists a wide-range of inferring techniques to infer the missing data of the uncovered subareas, such as expectation maximization (Schneider 2001) and singular spectrum analysis (Kondrashov and Ghil 2006). One of the powerful approach is the spatial-temporal compressive sensing (Kong et al. 2013; Zhang et al. 2009). The essential idea of this approach is based on the nonnegative matrix factorization (NMF) (Lee and Seung 2001; Mao and Saul 2004). Given the aggregated sensor data of recent sensing cycles (the number of recent sensing cycles used for NMF is denoted as $w$), this approach first sorts the subareas using their indices from $1 \dots$ to $|S|$, then maps the data into a $|S| \times w$ matrix denote as $R$, where the element $R_{a,t}$ ($1 \leq a \leq |S|$ and $1 \leq t \leq w$) refers to the aggregated sensing value of the $a^{th}$ subarea and $t^{th}$ sensing cycle (in the window). To recover the missing values in $R$, this approach obtains two non-negative **matrix factors** $P \in \mathbb{R}^{|S| \times l}$ and $Q \in \mathbb{R}^{l \times w}$ such that $R \approx PQ$, through NMF, where $l$ stands for the *Size of Latent Space* of NMF.

Typically, there are four key factors affecting the performance of the compressive community sensing: (1) *The Number of Subareas* that each participant covers in each sensing cycle; (2) *The Number of Participants (m)* which, together with the number of subareas per participant, can determine the coverage of collected sensor data; (3) *The Size of Windows (w)* that refers to the number of past sensing cycles used for matrix recovery (i.e., the width of the matrix for matrix completion); (4) *The Size of Latent Space (l)* that determines the

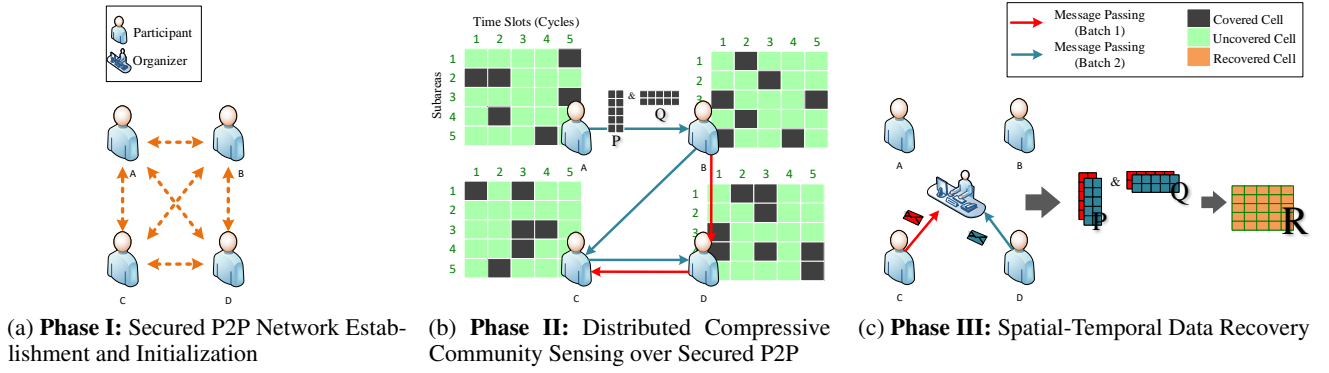

(a) **Phase I:** Secured P2P Network Establishment and Initialization

(b) **Phase II:** Distributed Compressive Community Sensing over Secured P2P

(c) **Phase III:** Spatial-Temporal Data Recovery

Figure 1: Overall Framework of $\mathcal{CSWA}$

rank of matrices for low-rank matrix recovery/completion.

## Problem Formulation

Given a set of participants, where each participant's mobile device stores the raw sensor data locally (without raw data sharing), our proposed work intends to recover the sensing data of the target area while assuming that the organizer is not allowed to aggregate the sensor data from any participants. Specifically, we make following assumptions:

- For all the sensing cycles in $T$ and subareas in $S$, there exists an unknown spatial-temporal sensor data matrix $R^*$ ($R^* \in \mathbb{R}^{|S| \times T}$), where each element $R^*_{a',t'}$ ($1 \leq a' \leq |S|$ and $1 \leq t' \leq |T|$) refers to the real value of sensor data in the corresponding subarea $a'$ and sensing cycle $t'$.

- In each sensing cycle (e.g., the $t^{th}$ cycle), each participant (e.g., the $j^{th}$ participant) covers a subset of subareas (i.e., $S_j^t \subseteq S$) in the target area. Thus, all the collected sensor data from the $1^{st}$ to the $t^{th}$ sensing cycle of the $j^{th}$ participant can be represented as a matrix $R^j \in \mathbb{R}^{|S| \times t}$, where each element refers to the value of the sensor data collected in the corresponding subarea and cycle. Note that, to protect the location privacy, $R^j$ is not known by the organizer.

- We denote the value of the sensor data collected by the $j^{th}$ participant in sensing cycle $t$ at subarea $a$ as $R_{a,t}^j$. Each sensor datum obtained is assumed to be the true value with (unknown) random noise, i.e., $R_{a,t}^j = R_{a,t}^* + \varepsilon_{a,t}^j$. For any two participants (i.e., the $j^{th}$ and $k^{th}$ participants), they might cover the same subarea (say, $a_j^t \cap S_k^t \neq \emptyset$ is possible), but are with *different* sensor data value obtained, due to the noise.

Our problem is that, in each sensing cycle $t$, with $R^j$ ($1 \leq j \leq N$) locally stored on each participant's device, there needs to estimate $\widehat{R}_{a,t}$ to

$$\textbf{minimize} \sum_{a=1}^{|S|} (\widehat{R}_{a,t} - R^*_{a,t})^2 \textbf{ for } 1 \leq t \leq T,$$

while ensuring that the organizer is prohibited to aggregate $R^j$ from any participant and the raw sensor/location data sharing is not allowed between the participants.

## Framework and Algorithms

In this section, we present the proposed framework of $\mathcal{CSWA}$ and the underlying algorithms. Specifically, we introduce a novel *Decentralized Spatial-Temporal Compressive Sensing* framework based on *Parallelized Stochastic Gradient Descent*.

### Framework Design

Before elaborating the proposed framework and algorithms, we make the following settings: (1) In order to simulate a secure peer-to-peer network over the community members, we define a set of participants, where these participants can receive or send messages (factor matrices) to each other trustfully and randomly; (2) When passing the message between two participants, the *receiver* can not send the updated matrix factors back to the *sender*, while the *sender* can easily recover the *receiver's* local sensing data by recalculating the return messages; (3) The organizer can only receive or access the related message when the updates (message passing) are finished. In this way, the private information such as real-time locations of the participants in each sensing cycle can be protected from the organizer.

The overall framework of $\mathcal{CSWA}$ consists of the following three phases (as illustrated in Figure.1):

### Phase I: Secure P2P Network Establishment and Initialization

Prior to initializing the *batch* on the organizer, we first establish a secure peer-to-peer (P2P) network among $m$ participants, while all the collected sensor data on the $j^{th}$ participant are mapped to a local data matrix $R^j$. Then, as shown in Algorithm. 1, $\mathcal{CSWA}$ randomly picks a set of participants which is the *batch* (denoted as the set $L$ with size $N$) from the secure network of $m$ participants. Next, given the target data matrix $R \in \mathbb{R}^{|S| \times w}$, $\mathcal{CSWA}$ extracts the row and column number of $R$ to construct the initial matrix factors $\hat{P}$ and $\hat{Q}$ on the organizer. Specifically, $\hat{P}_j$ is generated by a $|S| \times l$ Gaussian Random Matrix on the $j^{th}$ participant. Similarly, $\hat{Q}_j$ is generated by a $l \times w$ Gaussian Random Matrix on the same $j^{th}$ participant. To avoid the aforementioned message transferring back between two participants, we initialize a counter $i$ to record passing times (iterations) among participants and set $j_p$ to mark the last participant's index, where

**Algorithm 1: Initializing Batch and Matrix Factors $(\hat{P}, \hat{Q})$ on Organizer**

**Data:**
$R_{|S| \times w}$ — the target data matrix
**Parameter:**
/* Subareas covered by per participant */
$|S|$ — the maximum numbers of subareas
$w$ — the size of windows
$l$ — the size of latent space
**begin**
    /* Predefine a set of participants */
    **Randomly Draw** $N$ *Participants* into **Set** $L$
    /* $L = \{I_1, I_2, ..., I_N\}$ */
    **for** *each* $I_j \in L$ **do**
        /* Initialize matrix factors $P, Q$ on $I_j$ */
        $\hat{P}_j \leftarrow |S| \times l$ Gaussian Random Matrix
        $\hat{Q}_j \leftarrow l \times w$ Gaussian Random Matrix
        /* Initialize the counter and the previous participant index */
        **SEND** $(\hat{P}_j, \hat{Q}_j, \mathbf{0}, \mathbf{null})$ to **L**;
    **end**
**end**

the $(i, j_p)$ will be transferred along with the updated matrix factors so that the participant who receives the message can randomly select the next one excluding participant $j_p$. When the initialization ends, each participant $(I_j)$ in the predefined set $L$ (*batch*) will be assigned a pair of starting matrix factors $\hat{P}_j$ and $\hat{Q}_j$.

**Phase II: Distributed Compressive Community Sensing via Parallelized Low-Rank Approximation**

Given the mapped local data matrix $R^j$ on $j^{th}$ participant, $\mathcal{CSWA}$ intends to approximate the optimal estimation of matrix factors $\hat{P}_j$ and $\hat{Q}_j$ via parallelized stochastic gradient descent on top of non-negative matrix factorization algorithm. Specifically, the initialized $(\hat{P}_j, \hat{Q}_j, \mathbf{0}, \mathbf{null})$ has been allocated on the $j^{th}$ participant, where $\mathbf{0}$ refers to the fact that no update has been executed and "**null**" refers to there is no previous participant (coming from the organizer) which has updated the matrix factors (the index of previous participant is empty). Then the algorithm processes the updating task on each participant from the predefined *batch* ($L$) in parallel.

Suppose two dense matrix factors are $P \in \mathbb{R}^{|S| \times l}$ and $Q \in \mathbb{R}^{l \times w}$, the target minimization loss function over $m$ participants through parallelized stochastic gradient descent is as follow:

$$\hat{P}, \hat{Q} \leftarrow \underset{P \in \mathbb{R}^{|S| \times l}, Q \in \mathbb{R}^{l \times w}}{argmin} \left\{ \frac{1}{m} \sum_{j=1}^{m} \left\| F_j \circ (R^j - PQ) \right\|_F^2 + \lambda_P \|P\|_F^2 + \lambda_Q \|Q\|_F^2 \right\}, \quad (1)$$

where $l$ is the size of latent space, "$\circ$" means element-wise matrix multiplication, $\|\cdot\|_F$ is the Frobenius norm, $\lambda_P$ and $\lambda_Q$ are regularization parameters. Particularly, parallelly starting on each participant $I_j$, Algorithm. 2 first receives the input $(\hat{P}_j, \hat{Q}_j)$ from the last involved participant in the

**Algorithm 2: Parallelized Optimization on the $j^{th}$ Participant**

**Data:**
$R^j$ — the local data matrix on the $j^{th}$ participant
$F_j$ — the filter matrix on the $j^{th}$ participant
**Parameter:**
$i$ — the number iterations
$j_p, j$ — the index of previous and current participant
$\eta$ — step size
$\Delta_{min}$ — the minimum allowed perturbation
$t_{max}$ — the maximum number of allowed updates
$\lambda_P, \lambda_Q$ — regularization parameter on $P$ and $Q$ matrices
**begin**
    /* On receiving the message from the previous participant */
    **RECEIVE** $(\hat{P}_j, \hat{Q}_j, t, j_p)$
    /* Noting that "$A \circ B$" means element-wise matrix multiplication */
    $g_p \leftarrow (F_j \circ (R^j - \hat{P}_j \hat{Q}_j))\hat{Q}_j^T - \lambda_P \cdot \hat{P}_j$
    $g_q \leftarrow \hat{P}_j^T (F_j \circ (R^j - \hat{P}_j \hat{Q}_j)) - \lambda_Q \cdot \hat{Q}_j$
    $\hat{P}_j \leftarrow \hat{P}_j - \eta \cdot g_p$
    $\hat{Q}_j \leftarrow \hat{Q}_j - \eta \cdot g_q$
    /* Set the negative elements to zero */
    $\hat{P}_j, \hat{Q}_j \leftarrow Truncate(\hat{P}_j, \hat{Q}_j)$
    $i \leftarrow i + 1$
    /* Checking convergence conditions */
    $\Delta = max\{|g_p|_\infty, |g_q|_\infty\}$
    **if** $\Delta \geq \Delta_{max}$ **AND** $i \leq t_{max}$ **then**
        /* Not converged, continuing the algorithm */
        $j_{next} \leftarrow$ **Draw a random number from** $1$ **to** $m$ **except** $j_p$;
        **SEND** $(\hat{P}_j, \hat{Q}_j, i, j)$ **to the** $j_{next}^{th}$ **Participant**;
    **else**
        /* Converged, find out the optimal estimates */
        **SEND** $(\hat{P}_j, \hat{Q}_j)$ **to the Organizer**;
    **end**
**end**

secure network (or initialized from the organizer in the first run). Next it updates the $(\hat{P}_j, \hat{Q}_j)$ using the mapped local data matrix $R^j$ with the missing-value filter matrix $F_j$, and randomly picks up the next participant except the previous one ($j_p$) from the secure participants network and sends the updated $(\hat{P}_j, \hat{Q}_j)$ to this chosen participant. The matrix $F_j$ is a matrix filling with 0 (missing) and 1 (collected) which can set the missing elements in matrix $R^j$ to zero by the element-wise multiplication. We mainly use it to prevent the missing value in the local data matrix $R^j$ from affecting the gradient updating in $(\hat{P}_j, \hat{Q}_j)$. In addition, we leverage the *Truncate()* function, where the negative values in matrix factors $(\hat{P}_j, \hat{Q}_j)$ will be set to zero, then ensuring the non-negativeness of $(\hat{P}_j, \hat{Q}_j)$ when finishing each update.

Algorithm. 2 keeps picking up the next participant for updating, until the times of updates $i$ exceeds the maximal number of updates, or the updating process converges (i.e., $max\{|g_p|_\infty, |g_q|_\infty\} \leq \Delta_{max}$). Similar procedures are starting on each participant $I_j$ and the related matrix fac-tors keep updating independently. Once the updating process completes on each participant, Algorithm. 2 sends $(\hat{P}_j, \hat{Q}_j)$ where $j = 1, 2, ..., N$ to the organizer. When all the parallel processes are finished, the organizer has received $N$ pairs of the estimated $(\hat{P}, \hat{Q})$ for recovery of the target data matrix.

---

**Algorithm 3: Mobile Sensing Recovery on the Organizer**

**Data:**
$\hat{P}_j, \hat{Q}_j$ — the received matrix factors from the *batch*
**begin**
    /* Average all $\hat{P}_j, \hat{Q}_j$ on organizer     */
    $\bar{P} \leftarrow \frac{1}{N} \sum_{j=1}^{N} \hat{P}_j$
    $\bar{Q} \leftarrow \frac{1}{N} \sum_{j=1}^{N} \hat{Q}_j$
    /* Recover the target overall data matrix     */
    $\hat{R} \leftarrow \bar{P}\bar{Q}$
**end**

---

### Phase III: Spatial-Temporal Data Recovery

As we have introduced in the Preliminaries, the organizer can recover the target data matrix $\hat{R}$ based on the optimal estimated matrix factors $(\hat{P}, \hat{Q})$.

Given the received matrix factors $(\hat{P}_j, \hat{Q}_j)$ which are from the *batch*, Algorithm. 3 first separately average the $\hat{P}$ and $\hat{Q}$ from $j = 1$ to $N$. Then, to recover the target data matrix, the algorithm multiplies the averaged matrix factors $(\bar{P}, \bar{Q})$ and obtains the well-estimated target data matrix $\hat{R}$.

**Algorithm Analysis** In this section, we brief the analytical results of the proposed algorithms. Given the overall set of subareas $(S)$, the size of the latent space $(l)$, the size of the windows $(w)$, in each iteration, $N$ participants in the system would send out messages, while each participant sends a $|S| \times l$ matrix and a $l \times w$ matrix (i.e., $P$ and $Q$ matrices). In this way, the system-wide communication complexity in the worst-case (after the completion of $t_{max}$ iterations of message-passing) should be $\mathcal{O}\left((|S| \cdot l + l \cdot w) \cdot t_{max} \cdot N\right)$.

Suppose $P^*$ and $Q^*$ are the optimal solutions of the problem listed in Eq. 1, while $\bar{P}$ and $\bar{Q}$ (appeared in Algorithm 3) are two approximation results obtained by our algorithm. According to (Zinkevich et al. 2010), the approximation error of $||P^* - \bar{P}||_F \to 0$ and $||Q^* - \bar{Q}||_F \to 0$, when $t_{max \to +\infty}$ and $N$ is sufficiently large. Note that with a larger $N$, the proposed algorithm can achieve a faster rate of convergence of the approximation error with increasing $t_{max}$. For more theoretical analysis, please refer to (Zinkevich et al. 2010).

## Experiments

In order to evaluate the $\mathcal{CSWA}$ algorithm, we use the *Temperature (TEMP)* and *PM 2.5 air quality (PM25)* dataset, where the Experimental Setup section will cover all the settings and assumptions. Based on the above dataset, we first introduce the baseline algorithms which are commonly used in sensor data recovery. Specifically, the baseline algorithms adopt the ***matrix completion*** method and leverage the ***centralized computing*** patterns to recover the target sensing data. Then, we compare the performance of $\mathcal{CSWA}$ with baseline algorithms on two real-world datasets.

### Experimental Setup

For *TEMP* (Ingelrest et al. 2010) and *PM25* (Zheng, Liu, and Hsieh 2013) datasets, the sensing value of temperature (°C)/PM2.5 (air quality index) are located on each participant's mobile sensor in varying time slots (sensing cycle) and at different subareas. In details, the *TEMP* dataset contains the temperature readings in 57 cells (Subareas) and each sensing cycle lasts for 30 minutes. The *PM25* dataset includes the PM2.5 air quality values on 36 stations (Subareas) with the same sensing cycle.

In order to simulate the settings of the centralized computing patterns, we aggregate the collected sensing data from each participant. In details, we follow the aforementioned three phases to set the appropriate value of four key factors: *the Number of Participants ($m$), the Number of Subareas that each participant covers in each sensing cycle, the Size of Windows ($w$)* and *the Size of Latent Space ($l$)*. Note that each participant can sense the temperature/PM2.5 at a subset of subarea. Specifically, we use the maximum number of subareas $s$ ($1 \leq s \leq |S|$) in the experiments, assuming the participant can cover no more than $s$ subareas. To simulate the scenario that each participant can cover various number of subareas, the actual number of subareas covered by the participant will follow the discrete uniform distribution $U\{1, s\}$.

### Baseline Algorithms

In this section, we briefly introduce three baseline algorithms, where their advantages and drawbacks are listed as compared to $\mathcal{CSWA}$ algorithm.

- ***Spatio-Temporal Compressive Sensing (STCS)*** – STCS (Zhang et al. 2009; Wang et al. 2015) leverages the sparsity regularized matrix factorization to fill in the missing values in a certain matrix accounting for spatial-temporal properties. Based on the low-rank nature of real-world data matrices, STCS first exploits global and subarea structures in the data metrics. Then, it recovers the original matrices through matrix factorization under spatial-temporal constraints. Indeed, STCS advances ideas from compressive sensing and provides a highly effective (high accuracy and robustness) approach to solve the problem of missing data interpolation.

- ***Robust Principle Component Analysis (RPCA)*** and ***Truncated Singular Value Decomposition (TSVD)*** – RPCA (Gao et al. 2011) is derived from a widely used statistical procedure of principal component analysis (PCA), where RPCA performs well on solving the problem of matrices recovering. With respect to a mass of missing observations, RPCA aims to recover a low-rank matrix through random sampling techniques (Fischler and Bolles 1981). TSVD (Isam, Kanaras, and Darwazeh 2011) is also commonly used to approximate a low-rank matrix. Different from the traditional singular value decomposition, TSVD sets all but the first $k$ largest singular values equal

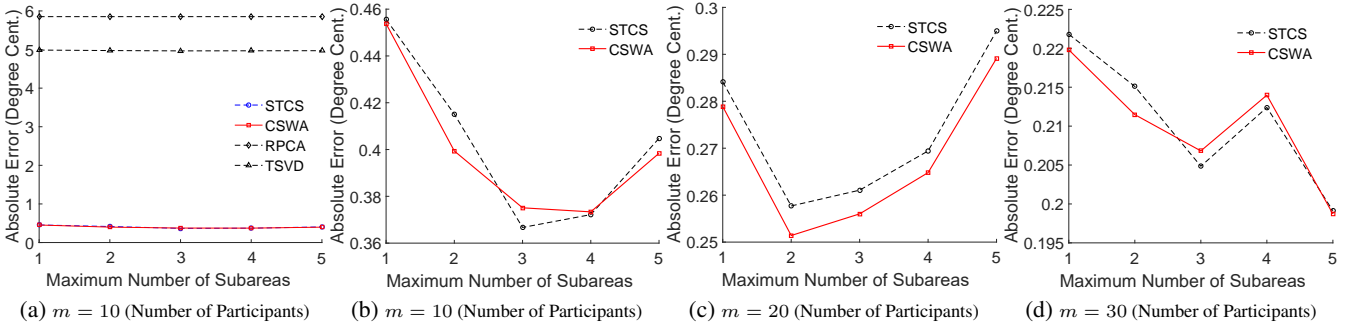

Figure 2: Performance Comparison with Varying *Maximum Number of Subareas* ($s$) per Participant per Cycle on *TEMP* Datasets.

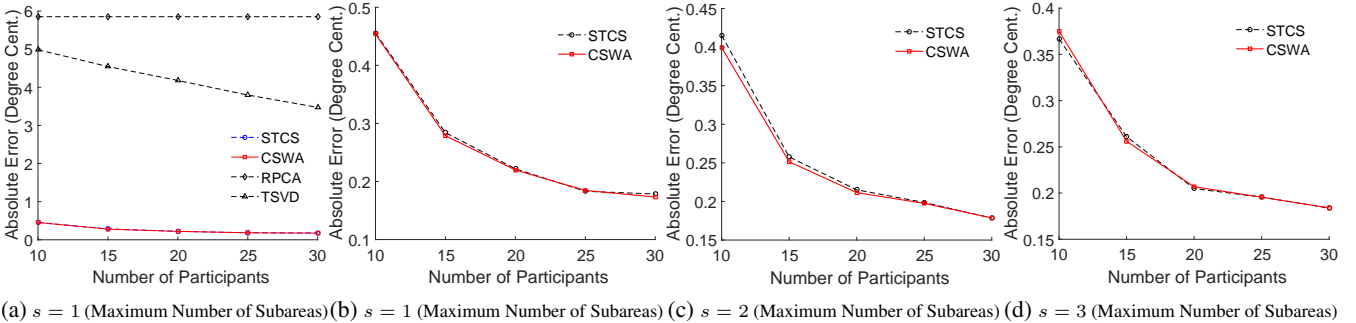

(a) $s = 1$ (Maximum Number of Subareas) (b) $s = 1$ (Maximum Number of Subareas) (c) $s = 2$ (Maximum Number of Subareas) (d) $s = 3$ (Maximum Number of Subareas) to zero and use only the first $k$ columns of the corresponding unitary matrices. With the optimality property, this method provides an efficient way to recover the target sensing matrix.

Figure 3: Performance Comparison with Varying *Number of Participants* ($m$) on *TEMP* Datasets.

## Experimental Results

In this section, we report the performance of $\mathcal{CSWA}$ and other three baselines on *TEMP* and *PM25* datasets. Specifically, we use the *Absolute Error*, which is the averaged element-wise difference $\left( \sum_{a=1}^{|S|} \sum_{t=1}^{|T|} |\widehat{R}_{a,t} - R^*_{a,t}|/(|S| \cdot |T|) \right)$ between the recovered matrix ($\widehat{R}$) and the original data matrix ($R^*$), as the indicator of the performance.

***TEMP* Datasets**. First, we present a comparison of algorithms with the settings of the maximum number of subareas (covered by each participant) ranging from $1$ to $5$ in Fig.2. Due to the overall better performances of $\mathcal{CSWA}$ and STCS, we present the entire comparison in (a) and only compare $\mathcal{CSWA}$ with STCS in other three settings (the same in Figures 3, 4 and 5 as well). Specifically, in Fig.2(a), 10 participants are involved. Then we vary the number of participants from $10$ to $30$ in the increment of $10$ in Figs.2(b), (c) and (d). We observe that the error is around $0.2$ to $0.45$ with varying maximum number of subareas from $1$ to $5$. It is noteworthy that $\mathcal{CSWA}$ can compete to STCS under these settings.

Second, we also compare $\mathcal{CSWA}$ with baseline algorithms by varying the number of participants in the secure P2P network. In Fig.3(a), the maximum number of subareas is 1. Then we increase it from $1$ to $3$ in the increment of $1$ in Figs.3(b), (c) and (d). In each comparison between $\mathcal{CSWA}$ and STCS, the error decreases when the number of participants increases for both of these two algorithms. This demonstrates that the larger group of participants can improve the performance of the matrix recovery, where intuitively the participants can cover more subareas and sensing cycles. Similar to the previous setting, $\mathcal{CSWA}$ can approximate the performance of STCS as well.

Further, we alter the values of two aforementioned key factors, such as *Size of Windows* and *Size of Latent Space*, to observe the variation of the error. Fig.4 shows that the error decreases when the window size increases from $20$ to $50$. Note that for each size of latent space in Figs.4(b), (c) and (d), the decreasing trends of the error are almost the same and the performance of $\mathcal{CSWA}$ still can compete with STCS. Fig.5 exhibits that the error increases when the size of the latent space increases from $2$ to $10$. Thus, for *TEMP* datasets, the small size of latent space can better approximate the original data matrix when it is low-rank. Thus the performance of $\mathcal{CSWA}$ is still competitive to STCS, as shown in Figs.5(b), (c) and (d).

***PM25* Datasets**. We conduct experiments with similar settings as *TEMP* datasets. Since the performances of RPCA and TSVD are still not as good as the other two algorithms, we only present the comparison between the proposed $\mathcal{CSWA}$ and STCS here. Specifically, in Table.1, we list the *Absolute Error* of these two algorithms with varying number of participants ($m$) and the window size ($w$). When the number of participants increases, the error is decreasing intuitively. On the contrary, the error increases with increased size of the window. However, $\mathcal{CSWA}$ performs comparably to STCS, sometimes even better (e.g., for $m = 20$). In Table.2, we show the performance with varying size of latent space and the number of subareas covered by each participant. The results reveal that the number of subareas does not affect the error significantly, while with the larger latent space the error is smaller with *PM25* datasets. Under these two settings, the performance of $\mathcal{CSWA}$ can still compete with STCS. Note that for each setting, we present the performance on the varying factor while keeping the other factor at optimal value. Also it is worth noting that the overall error is small on the average (10 with PM2.5 index ranging from 1 to 500) in both of $\mathcal{CSWA}$ and STCS.

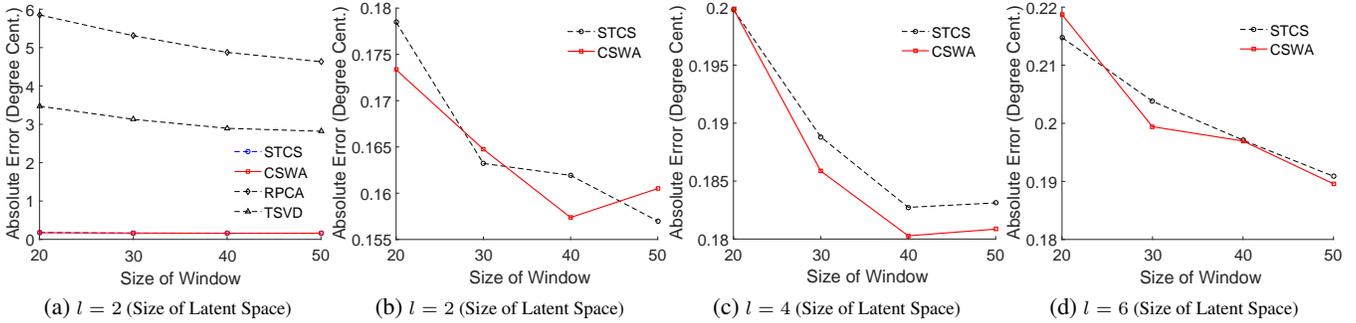

Figure 4: Performance Comparison with Varying *Size of Window* ($w$) on *TEMP* Datasets.

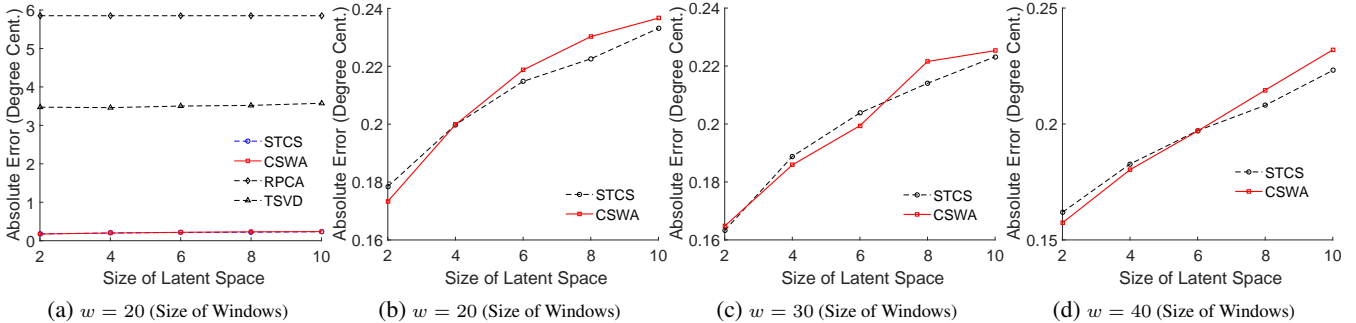

Figure 5: Performance Comparison with Varying *Size of Latent Space* ($l$) on *TEMP* Datasets.

Table 1: Performance Comparison (*Absolute Error*) with Varying *Number of Participants* ($m$) and *Size of Windows* ($w$) on *PM25* Datasets.

|      | Number of Participants ($m$) | | | Size of Windows ($w$) | | |
| --- | --- | --- | --- | --- | --- | --- |
|      | 10 | 20 | 30 | 20 | 30 | 40 |
| $\mathcal{CSWA}$ | 15.563 | 11.686 | 9.561 | 8.844 | 10.232 | 12.028 |
| STCS | 15.185 | 11.864 | 9.353 | 8.517 | 10.090 | 11.955 |

Table 2: Performance Comparison (*Absolute Error*) with Varying *Size of Latent Space* ($l$) and *Maximum Number of Subareas* ($s$) on *PM25* Datasets.

|      | Size of Latent Space ($l$) | | | Maximum Number of Subareas ($s$) | | |
| --- | --- | --- | --- | --- | --- | --- |
|      | 2 | 4 | 6 | 1 | 2 | 3 |
| $\mathcal{CSWA}$ | 11.777 | 9.561 | 8.945 | 8.166 | 8.945 | 8.844 |
| STCS | 11.518 | 9.353 | 8.719 | 8.220 | 8.719 | 8.516 |

**Summary:** With two real-world datasets, we compared the proposed $\mathcal{CSWA}$ with the baseline algorithms STCS, RPCA and TSVD. For both of the datasets, $\mathcal{CSWA}$ significantly outperforms RPCA and TSVD in most cases. Moreover, compared to the centralized algorithm STCS, $\mathcal{CSWA}$ also presents its competitiveness, with a low approximation error ($0.2°$ in city-wide temperature and 10 units of PM2.5 index in urban air quality). Even in some settings, the $\mathcal{CSWA}$ has a lower approximation error than STCS, which demonstrates the superiority of $\mathcal{CSWA}$.

## Conclusion

In this paper, we proposed $\mathcal{CSWA}$ – a novel community sensing paradigm. $\mathcal{CSWA}$ is designed to extract the environmental information in each subarea, without aggregating sensor and location data from the participants who partially cover the monitored area. On top of a secure peer-to-peer network over the participants, $\mathcal{CSWA}$ proposes a novel Decentralized Spatial-temporal Compressive Sensing framework based on Parallelized Stochastic Gradient Descent. Specifically, through learning the low-rank matrix structure via distributed optimization, $\mathcal{CSWA}$ approximates the value of sensor data in each subarea (both covered and uncovered) for each sensing cycle using the sensor data that locally stored in each participant's mobile device. According to the theoretical analysis on the parallelized stochastic gradient decent (Zinkevich et al. 2010), $\mathcal{CSWA}$ is capable of recovering the Spatial-Temporal information with bounded approximation error using the P2P communications of controllable complexity. The experiment results based on real-world datasets demonstrates that $\mathcal{CSWA}$ has low approximation error and performs comparably to (sometimes even better than) state-of-the-art algorithms based on the data aggregation and centralized computation.